%% file: main.tex
\documentclass[sigplan,10pt]{acmart}
\renewcommand\footnotetextcopyrightpermission[1]{}
\author{Zhangxiaowen Gong}
\affiliation{University of Illinois at Urbana-Champaign}
\email{gong15@illinois.edu}

\author{Houxiang Ji}
\affiliation{University of Illinois at Urbana-Champaign}
\email{hj14@illinois.edu}

\author{Christopher Fletcher}
\affiliation{University of Illinois at Urbana-Champaign}
\email{cwfletch@illinois.edu}

\author{Christopher Hughes}
\affiliation{Intel}
\email{christopher.j.hughes@intel.com}

\author{Josep Torrellas}
\affiliation{University of Illinois at Urbana-Champaign}
\email{torrella@illinois.edu}

\setcopyright{none}

\usepackage{booktabs}   

\usepackage[normalem]{ulem}
\usepackage{caption}
\usepackage{multirow}
\usepackage{color,soul}
\usepackage{subfig}

\usepackage[ruled,vlined,linesnumbered]{algorithm2e}

\newcommand{\name}{SparseTrain}

\begin{document}

%

\title{\textit{\name}: Leveraging Dynamic Sparsity in Training DNNs on General-Purpose SIMD Processors}

\setlength\tabcolsep{1.5pt}
%

%

%

\input{abstract}




\maketitle

\input{introduction}
\input{motivation}
\input{technique}

\input{expsetup}
\input{evaluation}
\input{relatedworks}
\input{conclusion}

%
\bibliographystyle{plain}
\bibliography{ref}

%

\end{document}

%% file: abstract.tex
\begin{abstract}

Our community has greatly improved the efficiency of deep learning applications, including by exploiting sparsity in inputs.  Most of that work, though, is for inference, where weight sparsity is known statically, and/or for specialized hardware.  We propose a scheme to leverage dynamic sparsity during training.  In particular, we exploit zeros introduced by the ReLU activation function to both feature maps and their gradients. This is challenging because the sparsity degree is moderate and the locations of zeros change over time.  We also rely purely on software.

We identify zeros in a dense data representation without transforming the data, and performs conventional vectorized computation. Variations of the scheme are applicable to all major components of training: forward propagation, backward propagation by inputs, and backward propagation by weights.  Our method significantly outperforms a highly-optimized dense direct convolution on several popular deep neural networks. At realistic sparsity, we speed up the training of the non-initial convolutional layers in VGG16, ResNet-34, ResNet-50, and Fixup ResNet-50 by 2.19x, 1.37x, 1.31x, and 1.51x respectively on an Intel Skylake-X CPU.

\end{abstract}

%% file: introduction.tex
\section{Introduction}\label{sec:intro}

Deep Neural Networks (DNNs) have become ubiquitous, achieving state-of-the-art results across a range of tasks from image recognition~\cite{Alexnet} to speech recognition~\cite{deep_speech_2} to scene generation~\cite{CNNGan} to game playing~\cite{AlphaGo}. While GPUs are amongst the fastest hardware solutions today for deep learning training, many institutions train on general-purpose processors. For example, in the supercomputing space, both Frontera~\cite{frontera} and SuperMUC-NG~\cite{leibniz-rechenzentrum}, the No. 5 and No. 9 supercomputers in the world respectively, as of June 2019, use only CPUs. In the datacenter space, companies such as Facebook have large datacenters with many CPUs, and use spare cycles of their CPUs to do training~\cite{takahashi2018}. Therefore, accelerating DNN training on general-purpose processors is an important yet sometimes undervalued area.

An effective approach to accelerating DNNs is to remove useless computations on zero values in the data, known as \emph{sparsity}.
Indeed, prior efforts spanning hardware to software to algorithms have exploited sparsity to eliminate computation or data transfers at different points in DNN computations. Most of these efforts, though, require hardware changes~\cite{Cnvlutin,Cambriconx,SCNN,EIE,Eyeriss,sparce,dma,recom} and/or apply only to \emph{inference}~\cite{Cnvlutin,scalpel,Cambriconx,HanPTD15,park2016faster,structured-sparsity-lasso,SCNN,EIE,Eyeriss}.  These are serious limitations.  Most real-world DNN computations are performed on conventional CPUs and GPUs~\cite{facebook_inference_datacenter,amazon_dnninference,facebook_inference_edge,nvidia_dnninference}, due to their wide availability, generality, and large memory capacity.  Further, significant time goes into \emph{training}.

This paper addresses these shortcomings through a \emph{software only} effort to speed up DNN training using sparsity, on unmodified general-purpose devices. 
This is challenging for multiple reasons.
First, works targeting sparse inference typically rely on sparse representations (e.g., Compressed Sparse Row, or CSR), which the sparsity pattern (i.e., locations of the non-zeros) is static~\cite{scalpel,Cambriconx,park2016faster,structured-sparsity-lasso,SCNN,EIE}.  For inference, the DNN weights are read-only, and so fit this criterion.  In training, though, the sparsity pattern in both the inputs and weights changes over time, since we update the weights with each batch of inputs.
Second, operating on sparse data incurs overhead.  Modern machines are highly optimized for dense computation, and suffer from extra indirections, branches, etc. in processing sparse data.  Prior work either relies on custom hardware to minimize these overheads~\cite{Cnvlutin,Cambriconx,SCNN,EIE,Eyeriss,sparce}, or sophisticated pre-processing to ``shape'' the sparsity pattern to better match existing hardware~\cite{park2016faster,scalpel,structured-sparsity-lasso} which, again, only applies for static sparsity.

Our method exploits the rectified linear unit (ReLU~\cite{Relu}), a ubiquitous operator used by convolutional neural networks (CNNs)~\cite{Alexnet,VGGnet,Googlenet,Resnet,Mobilenet,densenet}, multilayer perceptrons (MLPs)~\cite{TPU}, and recurrent neural networks (RNNs)~\cite{deep_speech_2}.
After each DNN \emph{layer}, all neurons (outputs) in the layer are passed through ReLU, which clamps each neuron's value to zero if it is negative. Whether a neuron is negative depends on the inputs and weights for that neuron, both of which change during execution. Thus, ReLU introduces \emph{dynamic sparsity}. However, ReLU usually only induces moderate sparsity, e.g., 40\%-90\%~\cite{dma}, compared to many scientific computing codes that exploit sparsity. Further, the sparsity pattern has no discernible structure. These factors make it difficult to overcome the overheads associated with exploiting sparsity.

We focus on CNNs. Given modest expected sparsity, {\em we leave data in a dense layout}, and exploit sparsity by detecting zero input values at runtime, and, when appropriate, branching over useless computations.  Our key observation is that in a CNN, each neuron has a large factor of reuse \emph{after} it passes through the ReLU; thus, with a good loop order, we can amortize the zero-detection and branching cost over lots of computation.  This is only the first step.  We introduce additional optimizations to minimize overhead while maximizing data locality, available parallelism, and the amount of work skipped per zero input. We name our scheme \emph{\name}.

\emph{\name} is general enough to be used on a variety of commercially available general-purpose processors.  However, some of our design decisions are influenced by an assumption of SIMD support.

We make the following contributions. First, we propose, to the best of our knowledge, the first DNN training algorithm that exploits sparsity from ReLU and applies to unmodified general-purpose devices. Second, we develop sparse methods that are decoupled from sparse representations and yield speedup at modest sparsity. Our algorithm is efficient even at processing dense input. Finally, our optimization techniques on register usage, reducing branch mispredictions, and data layout may provide insights to the community.


With profiled sparsity, we estimate that our method outperforms a highly optimized dense implementation by 2.19x for the non-initial convolutional layers in VGG16, 1.37x for ResNet-34, and 1.31x for ResNet-50, and 1.51x for the BatchNorm-free Fixup ResNet-50.

%% file: motivation.tex
\begin{table}[htbp]
  \centering
  \caption{List of the symbols and their dimensions/iterators}
  \label{tab:symbol}
  \footnotesize
  \begin{tabular}{p{0.033\linewidth}|p{0.24\linewidth}p{0.04\linewidth}||p{0.033\linewidth}|p{0.24\linewidth}p{0.16\linewidth}p{0.14\linewidth}}
    \toprule
     & description & itr. & & description & dim. & itr. \\
    \midrule
    $N$ & minibatch size & $i$ & $D$ & input tensor & $NCWH$ & $i,c,x,y$\\
    $C$ & input channels & $c$ & $Y$ & output tensor & $NKW'H'$ & $i,k,x',y'$\\
    $K$ & output channels & $k$ & $G$ & weight tensor & $KCRS$ & $k,c,u,v$\\
    $W$ & input width & $x$ & $L$ & loss function & &\\
    $H$ & input height & $y$ & $V$ & vector length & &\\
    $R$ & filter width & $u$ & $Q$ & $K$ tile size & &\\
    $S$ & filter height & $v$ & $T$ & \# skippable ops & &\\
    $O$ & horizontal stride & & $P$ & vertical stride & &\\
  \bottomrule
\end{tabular}
\end{table}

\section{Background}\label{sec:background}

\subsection{Training Convolutional Neural Networks}
\label{sec:convnet}

CNNs are a type of DNN that is effective for analyzing images. The leading competitors in recent years' ImageNet Large Scale Visual Recognition Competition (ILSVRC) are mostly variants of CNNs, such as AlexNet~\cite{Alexnet}, VGG~\cite{VGGnet}, GoogLeNet~\cite{Googlenet}, and ResNet~\cite{Resnet}. Within a CNN, the convolutional 
(i.e., {\em conv}) layers are the most time consuming components; thus, reducing the amount of computation in them can greatly boost performance.


The convolution on a minibatch of $N$ images with $C$ channels and size $H\times{}W$ correlates a set of $K$ filters with $C$ channels and size $S\times{}R$ on the images, producing a minibatch of $N$ images with $K$ channels and size $H/P\times{}W/O$, where $P$ and $O$ are the strides of the two dimensions, respectively. We denote filter elements as $G_{k,c,u,v}$ and image elements as $D_{i,c,x,y}$. The forward convolution for output $Y_{i,k,x',y'}$ is:
\begin{equation}
\small
\vspace{-1mm}
Y_{i,k,x',y'}=\sum^{C - 1}_{c=0}\sum^{R - 1}_{u=0}\sum^{S - 1}_{v=0}D_{i,c,x'\times{}O+u,y'\times{}P+v}\times{}G_{k,c,u,v}
\end{equation}

\begin{table*}[ht]
  \centering
  \caption{Evaluated layer configurations from VGG and ResNet v1.5}
  \label{tab:layers}
  \footnotesize
  \def\arraystretch{0.9}
  \begin{tabular}{p{0.08\linewidth}|p{0.04\linewidth}p{0.04\linewidth}p{0.03\linewidth}p{0.03\linewidth}p{0.014\linewidth}p{0.014\linewidth}p{0.014\linewidth}p{0.014\linewidth}||p{0.08\linewidth}|p{0.04\linewidth}p{0.04\linewidth}p{0.03\linewidth}p{0.03\linewidth}p{0.014\linewidth}p{0.014\linewidth}p{0.014\linewidth}p{0.014\linewidth}||p{0.08\linewidth}|p{0.04\linewidth}p{0.04\linewidth}p{0.03\linewidth}p{0.03\linewidth}p{0.014\linewidth}p{0.014\linewidth}p{0.014\linewidth}p{0.014\linewidth}}
    \toprule
    Name & $C$ & $K$ & $H$ & $W$ & $R$ & $S$ & $O$ & $P$ & Name & $C$ & $K$ & $H$ & $W$ & $R$ & $S$ & $O$ & $P$ & Name & $C$ & $K$ & $H$ & $W$ & $R$ & $S$ & $O$ & $P$\\
    \midrule
    vgg1\_2 & 64 & 64 & 224 & 224 & 3 & 3 & 1 & 1 & vgg2\_1 & 64 & 128 & 112 & 112 & 3 & 3 & 1 & 1 & vgg2\_2 & 128 & 128 & 112 & 112 & 3 & 3 & 1 & 1\\
    vgg3\_1 & 128 & 256 & 56 & 56 & 3 & 3 & 1 & 1 & vgg3\_2 & 256 & 256 & 56 & 56 & 3 & 3 & 1 & 1 & vgg4\_1 & 256 & 512 & 28 & 28 & 3 & 3 & 1 & 1\\
    vgg4\_2 & 512 & 512 & 28 & 28 & 3 & 3 & 1 & 1 & vgg5\_1 & 512 & 512 & 14 & 14 & 3 & 3 & 1 & 1 & resnet2\_1a & 64 & 64 & 56 & 56 & 1 & 1 & 1 & 1\\
    resnet2\_1b & 256 & 64 & 56 & 56 & 1 & 1 & 1 & 1 & resnet2\_2 & 64 & 64 & 56 & 56 & 3 & 3 & 1 & 1 & resnet2\_3 & 64 & 256 & 56 & 56 & 1 & 1 & 1 & 1\\
    resnet3\_1a & 256 & 128 & 56 & 56 & 1 & 1 & 1 & 1 & resnet3\_1b & 512 & 128 & 28 & 28 & 1 & 1 & 1 & 1 & resnet3\_2 & 128 & 128 & 28 & 28 & 3 & 3 & 1 & 1\\
    resnet3\_2/r & 128 & 128 & 56 & 56 & 3 & 3 & 2 & 2 & resnet3\_3 & 128 & 512 & 28 & 28 & 1 & 1 & 1 & 1 & resnet4\_1a & 512 & 256 & 28 & 28 & 1 & 1 & 1 & 1\\
    resnet4\_1b & 1024 & 256 & 14 & 14 & 1 & 1 & 1 & 1 & resnet4\_2 & 256 & 256 & 14 & 14 & 3 & 3 & 1 & 1 & resnet4\_2/r & 256 & 256 & 28 & 28 & 3 & 3 & 2 & 2\\
    resnet4\_3 & 256 & 1024 & 14 & 14 & 1 & 1 & 1 & 1 & resnet5\_1a & 1024 & 512 & 14 & 14 & 1 & 1 & 1 & 1 & resnet5\_1b & 2048 & 512 & 7 & 7 & 1 & 1 & 1 & 1\\
    resnet5\_2 & 512 & 512 & 7 & 7 & 3 & 3 & 1 & 1 & resnet5\_2/r & 512 & 512 & 14 & 14 & 3 & 3 & 2 & 2 & resnet5\_3 & 512 & 2048 & 7 & 7 & 1 & 1 & 1 & 1\\
  \bottomrule
\end{tabular}
\end{table*}

In the backward propagation of a conv layer, the gradient of the loss function $L$ w.r.t. the weights $G$ is calculated by applying the chain rule:
\begin{equation}
\small
\vspace{-1mm}
\frac{\partial L}{\partial G}=\frac{\partial L}{\partial Y}\frac{\partial Y}{\partial G}
\end{equation}
We need ${\partial L}/{\partial Y}$ from the next layer, and compute ${\partial L}/{\partial D}$ for the previous layer if needed. ${\partial L}/{\partial D}$ is a convolution of ${\partial L}/{\partial Y}$ with the layer's filters transposed. The gradient w.r.t. the weights is a convolution of $D$ with ${\partial L}/{\partial Y}$, producing $S\times{}R$ outputs for each input/output channel combination. 

Training a conv layer has three major components: the forward propagation (FWD), the backward propagation by input (BWI), and the backward propagation by weights (BWW). Table~\ref{tab:layers} lists the parameters of the layers that we evaluate.

\subsection{ReLU and Dynamic Sparsity}\label{sec:relu}

The output of a conv layer is usually passed through an activation function to introduce non-linearity. One popular activation is ReLU:
\begin{equation}
\small
    f(x)=max(0,x)
\end{equation}
and its derivative is\footnote{The derivative at $x=0$ is undefined but usually set to 0.}:
\begin{equation}
\small
    f'(x)= \bigg\{
  \begin{tabular}{ll}
  1, & if $x > 0$ \\
  0, & otherwise
  \end{tabular}
\end{equation}

By definition, ReLU and its derivative produce $50\%$ sparsity when the distribution of $x$ is centered at 0. When ReLU-activated conv layers are cascaded, this is reflected in $D$ in the forward propagation and ${\partial L}/{\partial Y}$ in the backward propagation, and it affects all three training components.

Since ReLU-induced sparsity varies with input, we call it {\em dynamic} sparsity to differentiate it from the {\em static} sparsity of weight-pruning. Dynamic sparsity is the only type that exists during the majority of the training time.\footnote{Static sparsity is also present when re-training a weight-pruned network, but we focus on regular dense training.}

Leveraging dynamic sparsity is challenging because the degree of sparsity is usually too low for a typical \emph{irregular} sparse computation to outperform highly optimized \emph{regular} dense computation.
In addition, at these modest sparsity levels, the metadata overheads of sparse representations such as CSR may exceed any savings. 


\subsection{Working Around Batch Normalization}\label{sec:batchnorm}

Batch normalization (BatchNorm)~\cite{ioffe2015batch} is a widely-adopted technique to facilitate training of deeper networks. BatchNorm first computes the mean and variance across the minibatch, normalizes the minibatch using those statistics, and then scales and shifts the normalized minibatch with learnable parameters.

In a CNN, BatchNorm is usually inserted between a conv layer and its subsequent ReLU. In that case, $\partial L/\partial Y$ of the conv layer no longer contains the sparsity produced by ReLU; thus, dynamic sparsity nearly disappears in BWI.

Fortunately, Zhang et al. showed that with proper initialization~\cite{zhang2019fixup}, one can train without BatchNorm with marginal accuracy loss. Removing BatchNorm restores the lost dynamic sparsity in BWI, and significantly accelerates training since BatchNorm take a notable portion of training time (24\% for ResNet-50~\cite{gitman2017comparison}).

\subsection{Baseline Platform}\label{sec:platform}

We consider a single node system comprising general-purpose processors with multiple cores and SIMD support.  While we tune and evaluate on a specific platform described in Section~\ref{sec:expsetup}, our approach is applicable to most modern shared-memory nodes with processors supporting SIMD.  

To provide context for our design decisions, we briefly describe our baseline platform.  We study a system with Intel Skylake cores. Each cycle, each core can execute up to two AVX-512 arithmetic instructions (e.g., fused multiply-add, or FMA), read two cache lines (64B) write one cache line from/to the L1 data cache, and retire a total of four instructions.  Each core has 32 vector registers, a 32KB L1 data cache, a 1MB L2 cache, and a 1.375MB non-inclusive L3 cache.

We leverage sparsity within a highly-tuned deep learning library, Intel's \emph{MKL-DNN}~\cite{mkl-dnn}. Our work is limited to generating additional convolution kernels for \emph{MKL-DNN} through the \emph{xbyak} just-in-time (JIT) assembler~\cite{xbyak}. Being low-level software, the implementation can easily and transparently be exploited at the application level, e.g., via deep learning frameworks like \emph{TensorFlow} or \emph{PyTorch}. 



%% file: technique.tex
\section{Exploiting Dynamic Sparsity}\label{sec:technique}

We leverage dynamic sparsity to speed up DNN training on shared-memory general-purpose SIMD processors. The idea is to skip operations that are rendered ineffectual by ReLU. Our scheme is called
\emph{\name}.

\emph{\name} uses a dense data representation for three reasons.
First, the
sparsity from ReLU is usually too low for any sparse representation 
to benefit. Second, we avoid the overhead of converting between dense and sparse representations. Finally, a dense format allows regular 
memory access patterns and more efficient vectorization.

In the following, we start by describing a na\"ive initial design, and then
progressively improve it.


\subsection{Na\"ive Forward Propagation}\label{sec:naivefwd}

We begin with direct convolution. Algorithm~\ref{al:basic} describes a na\"ive vectorized approach that reduces the operation count in FWD based on zeros in the input. Line~\ref{line:collapsed-0} and Line~\ref{line:collapsed-1} represent collapsed loop nests. For simplicity, the algorithm assumes unit stride, but can be easily changed for strided convolution. In the rest of the paper, we assume unit stride unless otherwise specified. The sparse algorithm for BWI is similar to FWD, and we will talk about BWW separately.


The main idea is, since an input element is reused $R\times{}S\times{}K$ times, by making the input stationary in the computation loop nest, we may skip at most $R\times{}S\times{}K$ calculations when we detect a zero.

We vectorize the computation along the output channel dimension ($K$). The statement in Line ~\ref{line:compute} represents a vector FMA operation of length $V$. When we detect a zero in Line ~\ref{line:judge}, we skip all of the following $R\times{}S\times{}K/V$ ineffectual FMAs. We denote the number of skippable FMAs per check as $T$. As shown in Table~\ref{tab:layers}, $K$ is often on the order of hundreds for later network layers. This, together
with the reuse of $R\times{}S$ means that, potentially, $T$ is large.




\begin{algorithm}
\small
\SetAlgoNoEnd
\SetAlgoNoLine
\SetInd{0.5em}{0.1em}
\SetKw{KwStep}{step}
\SetKw{KwIf}{if}
\SetKw{KwTrue}{is true}
\SetKwInOut{KwCon}{constant}
\SetKwInOut{Input}{input}
\SetKwInOut{Output}{output}
\Input{ input $D$, filters $G$}
\Output{ output $Y$}
 \For{$i=0,c=0,y=0,x=0$ \KwTo $N - 1,C - 1,H - 1,W - 1$}{\label{line:collapsed-0}
     \If{$D_{i,c,x,y} \neq 0$}{ \label{line:judge}
      \For{$k=0$ \KwTo $K - V$ \KwStep $V$}{
       \For{$u=0,v=0$ \KwTo $R - 1,S - 1$}{\label{line:collapsed-1}
         $Y_{i,[k:k+V-1],x-u,y-v} = Y_{i,[k:k+V-1],x-u,y-v} + D_{i,c,x,y}\times G_{[k:k+V-1],c,u,v}$\; \label{line:compute}
       }
      }
     }
 }
 \caption{Na\"ive Vectorized Sparse FWD}
 \label{al:basic}
\end{algorithm}


The na\"ive algorithm has several downsides. Firstly, it naturally has input parallelism: it compares each input element to zero, and then updates multiple output elements. Input parallelization requires atomic updates of the outputs, which drastically reduces performance. Output parallelization is generally faster. The simplest such approach is to let each core work on different images in the minibatch. However, common practice on training on CPU clusters is to assign a small minibatch to each node; thus, partitioning whole images may be too coarse grained, causing load imbalance.

Secondly, a CPU has a limited amount of architectural vector registers; this is 32 in the CPU we target. If $T=R\times{}S\times{}K/V$ is greater than the number of registers, we must spill registers during computation, inducing overhead. Therefore, we want to confine $T$. within the register budget.



Finally, the input's sparsity pattern is random, triggering branch mispredicts
in the zero-checking. Limiting $T$ to the register budget ($\sim$32), reduces our chance to amortize the misprediction penalty.


\subsection{Optimized Forward Propagation}\label{sec:optfwd}

This section introduces optimizations to improve the na\"ive FWD algorithm. Algorithm~\ref{al:outpar} shows the high-level ideas.

\begin{algorithm}
\small
\SetAlgoNoEnd
\SetAlgoNoLine
\SetInd{0.5em}{0.1em}
\SetKw{KwStep}{step}
\SetKw{KwIf}{if}
\SetKw{KwTrue}{is true}
\SetKw{KwPar}{in parallel}
\SetKw{KwFor}{for}
\SetKw{KwIn}{in}
\SetKwInOut{KwCon}{constant}
\SetKwInOut{Input}{input}
\SetKwInOut{Output}{output}
\Input{ input $D$, filters $G$}
\Output{ output $Y$}
 \For{$i=0$ \KwTo $N - M$ \KwStep $M$ \KwPar}{\label{line:2-firstloop-a}
  \For{$y=0$ \KwTo $H - 1$ \KwPar}{
   \For{$v=0$ \KwTo $S - 1$}{
    \For{$k=0$ \KwTo $K - Q$ \KwStep $Q$ \KwPar}{
     \For{$c=0$ \KwTo $C - V$ \KwStep $V$}{
      \For{$i'=i$ \KwTo $i + M - 1$ \KwPar}{\label{line:2-firstloop-b}
       \For{$x=0$ \KwTo $W - 1$}{\label{line:2-sweep}
        $m_{[0:V-1]}=$ [$d\neq 0$ \KwFor $d$ \KwIn $D_{i,[c:c+V-1],x,y+v}$]\;\label{line:2-loop}\label{line:2-check}
        \For{$c'=0$ \KwTo $V - 1$}{
         \If{$m_{c'}$ \KwTrue}{
          \For{$k'=k$ \KwTo $k+Q-V$ \KwStep $V$}{
           \For{$u=0$ \KwTo $R - 1$}{
            $Y_{i',[k':k'+V-1],x-u,y} = Y_{i',[k':k'+V-1],x-u,y} + D_{i',c+c',x,y+v} \times  G_{[k':k'+V-1],c+c',u,v}$\; \label{line:2-loop-e}
           }
          }
         }
        }
       }
      }
     }
    }
   }
  }
 }
 \caption{Parallel Vectorized Sparse FWD}
 \label{al:outpar}
\end{algorithm}


\subsubsection{Vectorized Zero-Checking}

The na\"ive algorithm compares input elements to zero one at a time. We vectorize this check along the input channel dimension ($C$). Line~\ref{line:2-check} does a vector comparison to generate a vector Boolean mask $m_{[0:V-1]}$; each mask bit indicates if the corresponding input element is zero. We then use the mask to control skipping computation.


\subsubsection{Increasing Output Parallelism}
\label{subsub_increase}

In a convolution, each input element affects a set of spatially grouped output elements. Similarly, any output element is calculated from a limited set of spatially grouped input elements. This allows us to increase output parallelism by reducing $T$.

We consider parallelizing at an output row granularity, similar to how \emph{MKL-DNN} parallelizes its direct convolution.  When a core works on an output row, it processes the input elements from $S$ corresponding input rows, one row at a time. This approach lowers $T$ from $R\times{}S\times{}K/V$ to $R\times{}K/V$. Moreover, when $R\times{}K/V$ is still larger than the number of registers, we further tile the output channel dimension ($K$) and decrease $T$ to $R\times{}Q/V$, where $Q$ is a factor of $K$ and a multiple of $V$. We will discuss how we choose $Q$ in the next section. We can process the same output row at different output channel tiles in parallel. With $T=R\times{}Q/V$, the number of parallel tasks rises from $N$ in the na\"ive algorithm to $N\times{}H\times{}K/Q$.


Since an input row corresponds to $S$ output rows, multiple cores may read a given input row. In a shared memory system, such reuse may be captured in a shared cache.


\subsubsection{Efficient Vector Register Usage}\label{sec:reg}

To avoid register spilling, we limit $T$. On the target CPU, the number of \texttt{zmm} vector registers is 32, and Algorithm~\ref{al:outpar} reserves a \texttt{zmm} register for holding the broadcasted input element $D_{i',c+c',x,y+v}$ in Line~\ref{line:2-loop-e} and keeps a vector of zeros for the vector compare instruction in Line~\ref{line:2-check}. Therefore, the register budget for $T$ is 30. On the target CPU, FMA instructions can take a memory operand, and the L1 read throughput matches the FMA throughput (2 per cycle per core); thus, we can operate on filter elements directly from memory.

We further reduce memory operations on output elements. As shown in 
Line~\ref{line:2-sweep} of Algorithm~\ref{al:outpar}, we scan through an input row and update the affected output elements accordingly. We call such a scan a {\em Row Sweep}. 
Due to a convolution's spatial nature, the outputs affected by adjacent input elements may overlap, depending on the filter width $R$ and the horizontal stride $O$. (Recall that we assume a unit stride in our discussion.) For example, when $R=3$ and $O=1$, $D_{i,c,x,y}$ affects $Y_{i,k,[x-2:x],y}$, and $D_{i,c,x+1,y}$ affects $Y_{i,k,[x-1:x+1],y}$. As a result, when we finish updating the output elements affected by $D_{i,c,x,y}$, we only need to save $Y_{i,k,x-2,y}$ to memory and load $Y_{i,k,x+1,y}$. On the other hand, $Y_{i,k,[x-1:x],y}$ can stay in registers. With this, each output element is only read and written once during a row sweep.

Moreover, since we JIT-generate kernels for different configurations, we can schedule the registers adequately according to $R$ and $O$ with a cyclic renaming scheme. In the above example, we may use \texttt{zmm[0:2]} as output buffers. When working on $Y_{i,k,[x-2:x],y}$, \texttt{zmm0} holds $Y_{i,k,x-2,y}$ while \texttt{zmm1} and \texttt{zmm2} hold $Y_{i,k,x-1,y}$ and $Y_{i,k,x,y}$ respectively. After moving on to $Y_{i,k,[x-1:x+1],y}$, \texttt{zmm0} proceeds to load $Y_{i,k,x+1,y}$ while $Y_{i,k,x-1,y}$ and $Y_{i,k,x,y}$ are kept in their previous registers. By keeping the renaming scheme consistent between the loads/stores and the FMAs, we avoid copying registers when moving from one input element to the next.

The cyclic renaming scheme requires unrolling the row sweep loop, starting on Line~\ref{line:2-sweep}. For large $W$, fully unrolling can lead to kernels too large for the instruction cache.  Since the cyclic renaming repeats every $R$ iterations, we instead unroll by a factor of $R$ to limit code size. 

Because $R$ and $V$ are fixed by the convolution configuration and the hardware, respectively, the only tunable parameter
in $T=R\times{}Q/V$ is $Q$. As a result, the register budget is often underutilized. 
To see why, assume we want $Q$ to be a factor of the number of output channels
$K$ so blocks have the same size. When $R=5$, $V=16$, and $K=256$, which is a typical number of channels, a reasonable maximum value of $Q$ is 64. As a result, $T=20$, leaving 10 registers unused. 

In such cases, we use spare registers to pipeline the load of output elements affected by the next input element. Again, using the above example and assuming
that we have a \texttt{zmm3} to spare, we can use it to load $Y_{i,k,x+1,y}$ while working on $Y_{i,k,[x-2:x],y}$, and schedule the cyclic renaming as if $R=4$. 
With this, FMAs depend on loads from an {\em earlier} iteration, and the out-of-order hardware can dispatch the FMAs sooner. When pipelining is enabled, the unroll factor of the row sweep loop becomes $R+1$ instead of $R$.

With pipelining, we use $(R+1)\times{}Q/V$ registers;
without it, we use $R\times{}Q/V$. We'd like this number to be as high as possible but no higher than the register budget. At $K=256$ and $V=16$, the optimal values of $Q$ for common values of the filter width $R$ are shown in Table~\ref{tab:regbuf}. As shown in the table, the values of $Q$ are 128 for $R=1$, 128 without pipelining for $R=3$, and 64 for $R=5$. For $R=1$, we found the alternative of $Q=256$ without pipelining is slower.


\begin{table}[htbp]
  \footnotesize
  \centering
  \caption{Optimal setup for $K=256$, $V=16$ at different $R$}
  \label{tab:regbuf}
  \begin{tabular}{c|cccc}
    \toprule
    $R$ & $Q$ & $T$ & Pipelined? & \# Registers \\
    \midrule
    1 & 128 & 8 & Y & 16\\
    3 & 128 & 24 & N & 24\\
    5 & 64 & 20 & Y & 24\\
  \bottomrule
\end{tabular}
\end{table}

\subsubsection{Reducing Branch Mispredictions}

As discussed, the optimal $T\leq 30$ on the target CPU. Under this constraint, the zero checking and skipping method in Algorithm~\ref{al:outpar} may induce so
many branch mispredictions that the code actually slows down. Therefore, we transform a series of branches to a single loop to reduce mispredictions.

Algorithm~\ref{al:branch} shows the 
method, and can replace Lines~\ref{line:2-check}-\ref{line:2-loop-e} in Algorithm~\ref{al:outpar}. First, we compare the input vector to zeros to generate a mask (maps to Line~\ref{line:2-check} in Algorithm~\ref{al:outpar}). This is done with the \texttt{vcmpps} instruction on the target CPU. Then, we use \texttt{popcnt} to count the number of 1s in the mask, which represents the number of non-zero elements in the input vector. After that, the code loops this number of times as shown in Line~\ref{line:3-loop2-b}-\ref{line:3-loop2-e} in Algorithm~\ref{al:branch}, where each loop iteration processes a non-zero element from the input vector.


\begin{algorithm}
\small
\SetAlgoNoLine
\SetInd{0.5em}{0.1em}
\SetKw{KwStep}{step}
\SetKwInOut{KwCon}{constant}
\SetKwInOut{Input}{input}
\SetKwInOut{Output}{output}
\Input{ input pointer \texttt{D}, filter pointer \texttt{G}}
\Output{ register array \texttt{Y}}
\KwCon{ filter offset \texttt{B}}
 \texttt{m[0:V-1] = vect\_cmp\_neq\_zero(D[0:V-1])}\; \label{line:3-vcmp}
 \texttt{o = population\_cnt(m[0:V-1])}\;
 \For{$i=0$ \KwTo $o - 1$}{ \label{line:3-loop2-b}
  \texttt{z = trailing\_zero\_cnt(m)}\;
  \texttt{D += z, G += z * B}\;
  \For{$j=0$ \KwTo $Q/V$}{ \label{line:3-loop1-b}
   \For{$k=0$ \KwTo $R$} {
    \texttt{Y[j][k][0:V-1]+=broadcast(D[0])*G[j][k][0:V-1]}
   }
  } \label{line:3-loop1-e}
  \texttt{m = shift\_right(m, z+1)}\;
  \texttt{D += 1, G += B}\;
 } \label{line:3-loop2-e}
 \caption{Zero Checking for Branch Performance; the loop nest at line~\ref{line:3-loop1-b}-\ref{line:3-loop1-e} is fully unrolled.}
 \label{al:branch}
\end{algorithm}

In each iteration, we first count the number of trailing zeros (\texttt{z}) in the mask with the \texttt{tzcnt} instruction.
Then, we advance the input pointer by \texttt{z}, to reach the next non-zero element in the input vector.
We also advance the filter pointer such that it points to the 
filter elements corresponding to the given non-zero input element.
Finally, we do the FMAs. 

We fully unroll the loop nest in Lines~\ref{line:3-loop1-b}-\ref{line:3-loop1-e}, and translate each \texttt{Y[j][k][0:V-1]} to a register name
(e.g., \texttt{zmm2}). In addition, the address calculation of 
\texttt{G[j][k][0:V-1]} depends on the shape of the filter array described in Section~\ref{sec:mem}. Finally, we
shift the mask to the right by \texttt{z}+1 to reflect that
we have finished processing the rightmost non-zero input element, and
also adjusts the input and filter pointers accordingly.

For readability, we omitted some low-level optimizations. Specifically, we pipeline the vector compare instruction such that the vector mask for the next iteration is generated during the current iteration. We also manually schedule and pipeline the integer instructions in the loop body to minimize dependence stalls. Moreover, we use shifts and load effective address (\texttt{lea}) instructions to reduce the strength of the integer multiplications and the number of integer instructions. In the end, each loop iteration of Lines~\ref{line:3-loop2-b}-\ref{line:3-loop2-e} contains 8 cheap integer instructions plus the FMAs. 

\subsubsection{Memory Access Optimization}\label{sec:mem}

We structured both the working sets and the loop nest carefully for high memory performance. First, we set the lowest dimension of the datasets to a channel tile of size $V$. On the target CPU, this is the \texttt{zmm} vector register size and the cache line size. Recall that we vectorize the computation along channels. Therefore, when the channel tile is aligned to a cache line boundary, vector instructions operate efficiently on a vector of the channel data.

We have 3 working sets, with different behaviors: the input $D$, the filters $G$, and the output $Y$.
$D$ and $Y$ have spatial locality during a row sweep. Each row element from them is loaded/stored only once per row sweep, and adjacent elements in a row are accessed consecutively. Such a streaming pattern benefits from hardware prefetching when we assign the second lowest dimension to the row dimension. We may also strategically software-prefetch the elements of the next row to the L2 cache when the line fill buffers (LFB) are not saturated.

In contrast, $G$ has temporal locality during a row sweep. Since we compute partial results for $W\times{}Q$ output elements from $W\times{}V$ input elements in a row sweep, we access $Q\times{}V\times{}R$ filter elements repeatedly. With the $R$ and $Q$ values listed in Table~\ref{tab:regbuf}, when $R=\{3,5\}$, 24KB or 20KB of $G$ elements are used per row sweep. Thus, on a machine with a 32KB L1-D cache, the next set of $G$ elements needs to be loaded from the L2 or below when the input/output channels of focus change. To counter the issue, we
block the minibatch dimension ($N$) with a tile size of $M$ to reuse each $G$ element $M$ times, as in Lines ~\ref{line:2-firstloop-a} and \ref{line:2-firstloop-b} in Algorithm~\ref{al:outpar}.  After testing, we confirmed that $M=16$ is appropriate for most convolution configurations.

We organize $G$ to leverage the hardware prefetcher. We set the lowest dimension to an output channel ($K$) vector of length $V$, the next dimension to an input channel ($C$) tile of length $V$, and the next dimension to the filter width dimension ($R$). When the kernel works on an input channel $c$ from a tile, it accesses $R\times{}Q/V$ output channel vectors. Thanks to the data layout, the hardware can prefetch the output channel vectors pertaining to $c+1$ in the meantime.

\subsection{Backward Propagation by Input}
\label{sec:backinput}

For a unit stride convolution, 
BWI is virtually the same as FWD, with the exception that the filters are transposed. However, non-unit strides introduce some differences. 
Specifically, when applying the register usage optimization described in Section~\ref{sec:reg} with row stride $O>1$, we load $Q/V$ new $Y$ element vectors into the register buffer after we finish processing $O$ vectors of $D$ during FWD. On the other hand, during BWI, we load $O\times{}Q/V$ new ${\partial L}/{\partial D}$ element vectors into the register buffers after we finish processing one ${\partial L}/{\partial Y}$ element vector.

Also, during a FWD row sweep, some $D$ elements may affect a number of $Y$ element vectors that is less than $T$ due to the horizontal stride; however, during a BWI row sweep, ignoring the image boundaries, an ${\partial L}/{\partial Y}$ element always affects $T$ ${\partial L}/{\partial D}$ element vectors. Our JIT based implementation can correctly generate the appropriate number of skippable FMAs accordingly.

Finally, the unroll factor of the row sweep loop in FWD is $W\times{}O$. In BWI, it is the least common multiple of $W$ and $O$.

\subsection{Backward Propagation by Weights}
\label{sec:backweight}

Algorithm~\ref{al:basicbww} is a na\"ive sparse algorithm for BWW. It checks for zeros in $D$. We can easily modify the algorithm to check for zeros in ${\partial L}/{\partial Y}$ instead, if we expect more sparsity in ${\partial L}/{\partial Y}$ of the target layer. In Algorithm~\ref{al:outparbww}, we apply output-parallelization and similar optimizations used in the other two components, with some changes.

\begin{algorithm}
\small
\SetAlgoNoEnd
\SetAlgoNoLine
\SetInd{0.5em}{0.1em}
\SetKw{KwStep}{step}
\SetKw{KwIf}{if}
\SetKw{KwTrue}{is true}
\SetKwInOut{KwCon}{constant}
\SetKwInOut{Input}{input}
\SetKwInOut{Output}{output}
\Input{ input $D$, output gradients $dY$}
\Output{ filter gradients $dG$}
 \For{$i=0,c=0,y=0,x=0$ \KwTo $N - 1,C - 1,H - 1,W - 1$}{
     \If{$D_{i,c,x,y} \neq 0$}{
      \For{$k=0$ \KwTo $K - V$ \KwStep $V$}{
       \For{$u=0,v=0$ \KwTo $R - 1,S - 1$}{
         $dG_{[k:k+V-1],c,u,v} = dG_{[k:k+V-1],c,u,v} + D_{i,c,x,y}\times dY_{i,[k:k+V-1],x-u,y-v}$\;
       }
      }
     }
 }
 \caption{Na\"ive Vectorized Sparse BWW}
 \label{al:basicbww}
\end{algorithm}

\begin{algorithm}
\small
\SetAlgoNoEnd
\SetAlgoNoLine
\SetInd{0.5em}{0.1em}
\SetKw{KwStep}{step}
\SetKw{KwIf}{if}
\SetKw{KwTrue}{is true}
\SetKw{KwPar}{in parallel}
\SetKwInOut{KwCon}{constant}
\SetKw{KwFor}{for}
\SetKw{KwIn}{in}
\SetKwInOut{Input}{input}
\SetKwInOut{Output}{output}
\Input{ input $D$, output gradients $dY$}
\Output{ filter gradients $dG$}
 \For{$i=0$ \KwTo $N - V$ \KwStep $V$}{
  \For{$y=0$ \KwTo $H - 1$}{
   \For{$v=0$ \KwTo $S - 1$ \KwPar}{
    \For{$k=0$ \KwTo $K - Q$ \KwStep $Q$ \KwPar}{
     \For{$c=0$ \KwTo $C - 1$ \KwPar}{
      \For{$x=0$ \KwTo $W - 1$}{
       $m_{[0:V-1]}=$ [$d \neq 0$ \KwFor $d$ \KwIn $D_{[i:i+V-1],c,x,y+v}$]\;\label{line:5-check}
       \For{$i'=0$ \KwTo $V - 1$}{
        \If{$m_{i'}$ \KwTrue}{
         \For{$k'=k$ \KwTo $k+Q-V$ \KwStep $V$}{\label{line:5-loop}
          \For{$u=0$ \KwTo $R - 1$}{
           $dG_{[k':k'+V-1],c,u,v} = dG_{[k':k'+V-1],c,u,v} + D_{i+i',c,x,y+v}\times dY_{i+i',[k':k'+V-1],x-u,y}$\;
          }
         }
        }
       }
      }
     }
    }
   }
  }
 }
 \caption{Parallel Vectorized Sparse BWW}
 \label{al:outparbww}
\end{algorithm}

We vectorize the zero-checking along the minibatch dimension ($N$) instead of the channel dimension as in FWD and BWI, reflected in Line~\ref{line:5-check}, because in BWW, the destination of the FMA operation, $dG_{[k:k+V-1],c,u,v}$, changes as the input channel $c$ changes. As a result, if we vectorize the zero-checking along the input channel dimension ($C$), we need to store the previous group of $dG_{[k:k+V-1],c,u,v}$ vector to memory and load a new group before entering the loop starting at Line~\ref{line:5-loop}, and this frequent register spilling may harm performance significantly. Luckily, because $dG_{[k:k+V-1],c,u,v}$ is minibatch-invariant, all input elements from the vector $D_{[i:i+V-1],c,x,y+v}$ affects the same group of $dG_{[k:k+V-1],c,u,v}$. Therefore, vectorizing the zero-checking along the minibatch dimension avoids spilling the registers.

Due to the change in vectorization scheme, we transpose the input $D$ such that the lowest dimension is a minibatch tile of size $V$. This is an effort to avoid gathering.

In a row sweep, a core works on $R\times{}Q$ filter gradient elements. Because the total number of filter gradient elements is $R\times{}S\times{}K\times{}C$, the maximum parallelism becomes $S\times{}C\times{}K/Q$.

Since the set of filter gradient elements is constant during a row sweep, if we limit the number of filter gradient vectors being worked on, which is $T=R\times{}Q/V$, to the register budget, they can stay in the registers during the entire row sweep. Consequently, we do not apply the cyclic register load/store and renaming scheme described in Section~\ref{sec:reg}. This also lifts the restriction on the unrolling factor for the row sweep loop so that it can be chosen freely.

Instead of loading the previous partial results of the filter gradient vectors at the beginning of a row sweep and store the new partial results to memory at the end, we clear the output buffer registers at the beginning and store the FMA results in them during a row sweep. At the end, we load the previous partial results and add them to the output buffer registers as the new partial results, and we immediately store them back to memory afterwards. Therefore, the filter gradient elements are only accessed twice in succession at the end. We also prefetch the filter gradient elements in software at the beginning. With the optimization, tiling the minibatch dimension to reuse the filter elements as described in Section~\ref{sec:mem} is unnecessary.  

The two source operands of the FMA instructions used in BWW are the broadcasted input element $D_{i+i',c,x,y}$ in a \texttt{zmm} register and the ${\partial L}/{\partial Y}$ vector $dY_{i+i',[k':k'+V-1],x-u,y}$ as a memory operand.

%% file: expsetup.tex
\section{Experimental Setup}\label{sec:expsetup}

We build \emph{\name} as additional convolution kernels in \emph{MKL-DNN}, and use the \emph{xbyak} JIT assembler to generate the code. We use \emph{MKL-DNN} v0.90's direct convolution kernel as the baseline, refered to as \emph{direct}. Georganas et al.~\cite{georganas2018anatomy} documented most of the optimizations employed by the baseline.

We compare the performance of \emph{\name} against \emph{MKL-DNN}'s on an Intel Core i7-7800X Skylake-X CPU with 6 cores, 2 AVX-512 vector units per core, a 32KB L1 D-cache and 1MB L2 cache per core, and 8.25MB of shared L3 cache. We disable hyperthreading as well as dynamic frequency scaling and enable 2MB pages.

Because our kernels are JIT generated, the choice of compiler does not affect our performance much, but it may impact some of \emph{MKL-DNN}'s implementations. We use the Intel C++ Compiler (\emph{ICC}) 19.0 for the experiments.

To evaluate \emph{\name} at various sparsity levels, we generate synthetic input with random sparse patterns and experiment on all but the first conv layers from VGG~\cite{vgg} and ResNet~\cite{Resnet}. We use a batch size of 16 during the experiment. Table~\ref{tab:layers} lists the experimented layer configurations.

Finally, we estimate the speedup in the end-to-end training of the conv layers of VGG16, ResNet-34/50, and a variant of the Fixup ResNet-50~\cite{zhang2019fixup}. The original Fixup ResNet eliminates BatchNorm but adds a scalar bias between each ReLU and the next conv layer, which erases dynamic sparsity in FWD. We remove those bias terms with a 1.0\% penalty in top-1 accuracy.

For the three ResNet variants, we randomly select 5 minibatches and profile their real sparsity patterns throughout 100-epoch ImageNet training sessions. We then run \emph{\name} against the profiled sparsity patterns to project the total execution time of the conv layers in the whole training. For VGG16, we use the sparsity levels profiled by Rhu et al.~\cite{dma}, and we generate synthetic input at the profiled sparsity levels to project the total execution time.


%% file: evaluation.tex
\section{Evaluation}\label{sec:evaluation}

\subsection{$3\times{}3$ Convolutional Layers}


\begin{figure}[htbp]
    \centering
    \includegraphics[width=\linewidth]{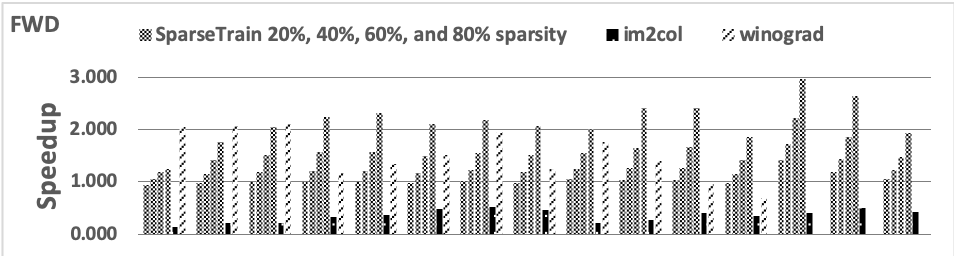}\\
    \includegraphics[width=\linewidth]{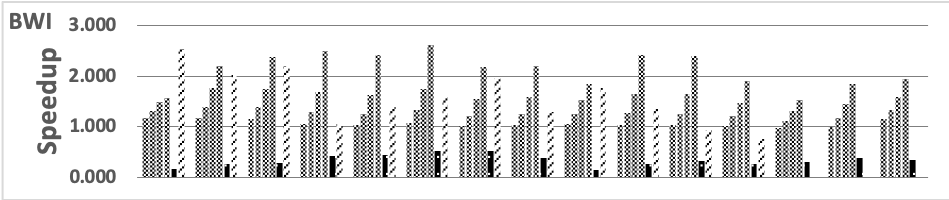}\\
    \includegraphics[width=\linewidth]{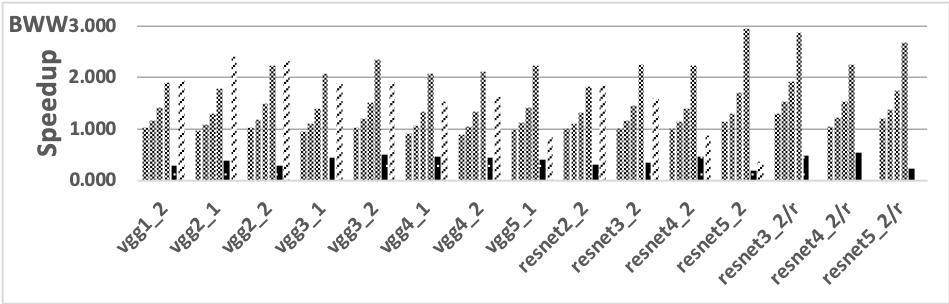}\\
    \caption{The speedup over \emph{direct} on the 3x3 layers}
    \label{fig:3x3}
\end{figure}

$3\times{}3$ ($R=S=3$) has become the most popular convolutional layer type in recent years, so the performance of them is crucial. Techniques such as the \emph{Winograd} algorithm~\cite{Winograd} have been proposed to accelerate $3\times{}3$ Layers, and \emph{MKL-DNN} implements a highly optimized vectorized \emph{Winograd} convolution that often outperforms \emph{direct}. However, because the \emph{Winograd} algorithm reduces computation by transforming the problem to the ``Winograd space,'' it has two drawbacks that are absent in \emph{\name}. First, the transformation introduces numerical instability as the filter size increases, so its application is usually limited to $3\times{}3$ Layers~\cite{winostability}; second, it requires additional workspace memory. Further, \emph{MKL-DNN}'s Winograd implementation does not support strided convolution.

Besides the aforementioned algorithms, \emph{MKL-DNN} also implements a \emph{im2col} based convolution. The algorithm flattens and duplicates parts of the input image and the filters to form matrices and then performs matrix multiplication with \emph{gemm} calls. The version of \emph{MKL-DNN} that we use incorporates \emph{MKL} 2019.0 as the backend \emph{gemm}. Although \emph{gemm} itself is highly optimized, creating the matrices incurs time and memory overheads, so this implementation is generally slower than \emph{direct}. Figure~\ref{fig:3x3} shows the speedup of \emph{\name} at 20-80\% sparsity over \emph{direct} for all three training components of studied layers. We compare against \emph{im2col} and \emph{Winograd} when applicable. Table~\ref{tab:3x3speedup} lists the geo-mean speedup at various sparsity.

At $0\%$ sparsity (i.e., a truly dense input), \emph{\name} reaches $92\%$-$95\%$ of \emph{direct}'s performance on average, depending on the component. This indicates that the overhead to check for and exploit sparsity is low, and the loop order as well as the tiling strategy of \emph{\name} are effective.

\begin{table}[htbp]
  \caption{Average speedup at different sparsity for $3\times{}3$ layers}
  \label{tab:3x3speedup}
  \footnotesize
  \begin{tabular}{p{0.085\linewidth}|p{0.06\linewidth}p{0.06\linewidth}p{0.06\linewidth}p{0.06\linewidth}p{0.06\linewidth}p{0.06\linewidth}p{0.06\linewidth}p{0.06\linewidth}p{0.06\linewidth}p{0.07\linewidth}|p{0.07\linewidth}|p{0.06\linewidth}}
    \toprule
     & \multicolumn{10}{c|}{\emph{\name}} & \multirow{2}{*}{\emph{im2c.}} & \multirow{2}{*}{\emph{win.}}\\
     & $0\%$ & $10\%$ & $20\%$ & $30\%$ & $40\%$ & $50\%$ & $60\%$ & $70\%$ & $80\%$ & $90\%$ & &\\
    \midrule
    FWD & 0.92 & 0.96 & 1.04 & 1.13 & 1.24 & 1.38 & 1.56 & 1.79 & 2.11 & 2.48 & 0.33 & 1.45\\
    BWI & 0.93 & 0.98 & 1.06 & 1.15 & 1.26 & 1.40 & 1.58 & 1.81 & 2.10 & 2.45 & 0.31 & 1.48\\
    BWW & 0.95 & 0.98 & 1.03 & 1.10 & 1.18 & 1.30 & 1.48 & 1.76 & 2.23 & 3.15 & 0.37 & 1.44\\
  \bottomrule
\end{tabular}
\end{table}

On average, the sparsity cross-over point for \emph{\name} to outperform \emph{direct} is between $10\%$-$20\%$, which is lower than the realistic sparsity during training. At $50\%$ sparsity, which is the expected value at the beginning of the training when the distribution of the weights is centered at 0, \emph{\name} on average delivers a 1.30x-1.40x speedup.

Typically, the later layers in a network have higher sparsity than the earlier layers. The sparsity reaches over $90\%$ for VGG16 and ResNet-34 layers, and over $80\%$ for ResNet-50 layers. At such level, \emph{\name} is on average over 2x faster than \emph{direct}. On the contrary, \emph{im2col} is always significantly slower than the baseline. When the stride is 1, \emph{Winograd} is on average 1.44x-1.48x faster than \emph{direct}.

\emph{\name} performs better at later layers while \emph{Winograd} dominates at earlier layers. This is partly due to the increased sparsity at later layers; on average, it takes at least $50\%$-$60\%$ sparsity for \emph{\name} to surpass \emph{Winograd}.  The other factor is a smaller number of channels for earlier layers, which limits the number of skippable FMAs per input element, and thus reduces efficiency. For example, both \emph{vgg1\_2} and \emph{resnet2\_2} have $C$ and $K$ of 64, giving us only 12 skippable FMAs.  Since \emph{\name} and \emph{Winograd} have different specialties, they can supplement each other.


With stride 1, \emph{\name} for FWD and BWI have similar performance.  However, for stride-2 layers (\emph{resnet3\_2/r}, \emph{resnet4\_2/r}, and \emph{resnet5\_2/r}), the former outperforms the latter. As discussed in Section~\ref{sec:backinput}, $\partial L/\partial D$ needs to be loaded $O^2$ times more rapidly during a row sweep in BWI than $Y$ being loaded in FWD. Therefore, BWI suffers from cache bandwidth limitations.

\subsection{$1\times{}1$ Convolutional Layers}


\begin{figure}[ht]
    \centering
    \includegraphics[width=\linewidth]{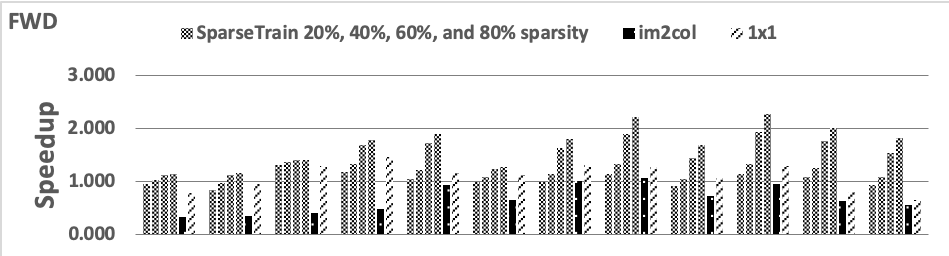}\\
    \includegraphics[width=\linewidth]{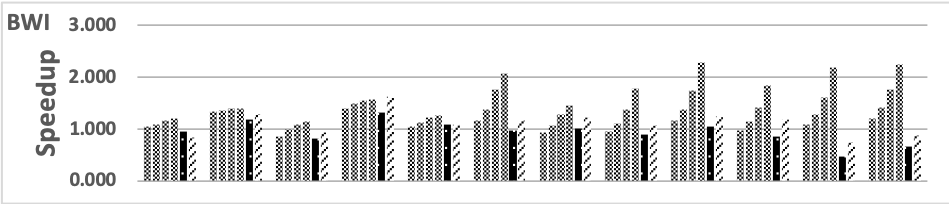}\\
    \includegraphics[width=\linewidth]{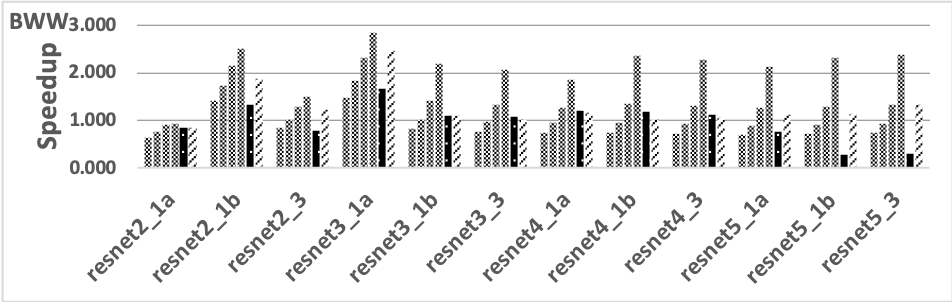}
    \caption{The speedup over \emph{direct} on the 1x1 layers}
    \label{fig:1x1}
\end{figure}

$1\times{}1$ layers ($R=S=1$) are widely used in ResNet-50's bottleneck blocks. They are unique amongst convolutions in that the spatial reuse of $R\times{}S$ is completely absent. As a result, an output element is just a weighted sum of all input channels at the corresponding input $x,y$ location. \emph{MKL-DNN} provides a specialized algorithm that uses a reduction instead of the accumulation employed by the baseline to specifically deal with $1\times{}1$ Layers. We call it the \emph{1x1} kernel.

Figure~\ref{fig:1x1} shows the speedup on each $1\times{}1$ layer over the dense direct from \emph{\name}, \emph{im2col}, and \emph{1x1}. Table~\ref{tab:1x1speedup} lists the average speedup at different sparsity. \emph{\name} is developed under the premise that convolution has a high compute-to-memory ratio. However, the ratio for $1\times{}1$ layers is 9x lower than that for $3\times{}3$ layers with the same input/output/channel sizes; thus, as we eliminate useless FMAs, $1\times{}1$ layers may become bandwidth-bound sooner than $3\times{}3$ layers. Therefore, at high sparsity, \emph{\name} is less effective on $1\times{}1$ layers than on $3\times{}3$ layers, only reaching 1.66x-2.04x speedup on average at $80\%$ sparsity.

\begin{table}[htbp]
  \caption{Average speedup at different sparsity for $1\times{}1$ layers}
  \label{tab:1x1speedup}
  \footnotesize
  \begin{tabular}{l|p{0.06\linewidth}p{0.06\linewidth}p{0.06\linewidth}p{0.06\linewidth}p{0.06\linewidth}p{0.06\linewidth}p{0.06\linewidth}p{0.06\linewidth}p{0.06\linewidth}p{0.07\linewidth}|p{0.07\linewidth}|p{0.06\linewidth}}
    \toprule
     & \multicolumn{10}{c|}{\emph{\name}} & \multirow{2}{*}{\emph{im2c.}} & \multirow{2}{*}{\emph{1x1}}\\
     & $0\%$ & $10\%$ & $20\%$ & $30\%$ & $40\%$ & $50\%$ & $60\%$ & $70\%$ & $80\%$ & $90\%$ & &\\
    \midrule
    FWD & 0.97 & 0.98 & 1.03 & 1.09 & 1.17 & 1.27 & 1.39 & 1.51 & 1.66 & 1.78 & 0.62 & 1.06\\
    BWI & 1.03 & 1.03 & 1.08 & 1.15 & 1.22 & 1.33 & 1.43 & 1.53 & 1.66 & 1.76 & 0.91 & 1.08\\
    BWW & 0.71 & 0.76 & 0.83 & 0.92 & 1.05 & 1.20 & 1.39 & 1.66 & 2.04 & 2.61 & 0.87 & 1.23\\
  \bottomrule
\end{tabular}
\end{table}

We also notice that BWW behaves differently than the other two components. At $0\%$ sparsity, \emph{\name}'s performance is on par with the baseline for FWD and BWI. For BWW, though, \emph{\name} only reaches $71\%$ of baseline. However, at high sparsity, \emph{\name}'s speedup is higher for BWW than the other two components.

Here we compare BWW with FWD. Its difference with BWI can be derived. The difference stems from two competing factors both related to how BWW accesses $\partial L/\partial Y$ against how FWD accesses $Y$. First, BWW uses a different loop order, and in a row sweep touches $V$ times more elements from $\partial L/\partial Y$ than FWD touches $Y$ at $0\%$ sparsity. Second, BWW reads $\partial L/\partial Y$ elements as a memory operand of an FMA. When we skip a group of FMAs, we also skip the access to the $\partial L/\partial Y$ elements.  At high sparsity, we eliminate many such access. In contrast, FWD loads and stores $Y$ elements using the cyclic register allocation scheme described in Section~\ref{sec:reg}, so the $Y$ elements are loaded and stored regardless of sparsity pattern.  Therefore, at low sparsity, BWW performs many more memory accesses, and at high sparsity, performs many fewer. The effect of the above factors is less visible at $3\times{}3$ layers thanks to their higher compute-to-memory ratio; however, it surfaces at $1\times{}1$ layers.



The lower channel sizes at earlier $1\times{}1$ layers hurts \emph{\name} more than they do at earlier $3\times{}3$ layers due to the absence of spatial reuse. For example, \emph{resnet2\_1a} has 64 for $C$ and $K$, resulting in only 4 FMAs being skippable per zero-checking. Consequently, we can hardly see speedup from \emph{\name} on earlier $1\times{}1$ layers. Nonetheless, we can still efficiently leverage the dynamic sparsity in later $1\times{}1$ layers.

On average, the cross-point sparsity for \emph{\name} to surpass the specialized \emph{1x1} kernel is below $30\%$  for FWD as well as BWI, and around $50\%$ for BWW.

In addition to $1\times{}1$ and $3\times{}3$ layers, we also experimented with several $5\times{}5$ layers and got even higher speedup. We omit the results due to lack of popularity of the $5\times{}5$ layers.

\subsection{Performance at Profiled Sparsity}\label{sec:finalresults}


Rhu et al.~\cite{dma} observed that during training, the sparsity from ReLU often begins at $\sim$50\% but increases rapidly in the first several epoches, and then slowly decreases. Also, later conv layers generally have higher sparsity then earlier layers. They further demonstrated that most of VGG16's layers are over 80\% sparse on average, and some layers' outputs may reach $90\%$ sparsity on average.

\begin{figure}[htbp]
    \centering
    \includegraphics[width=1\linewidth]{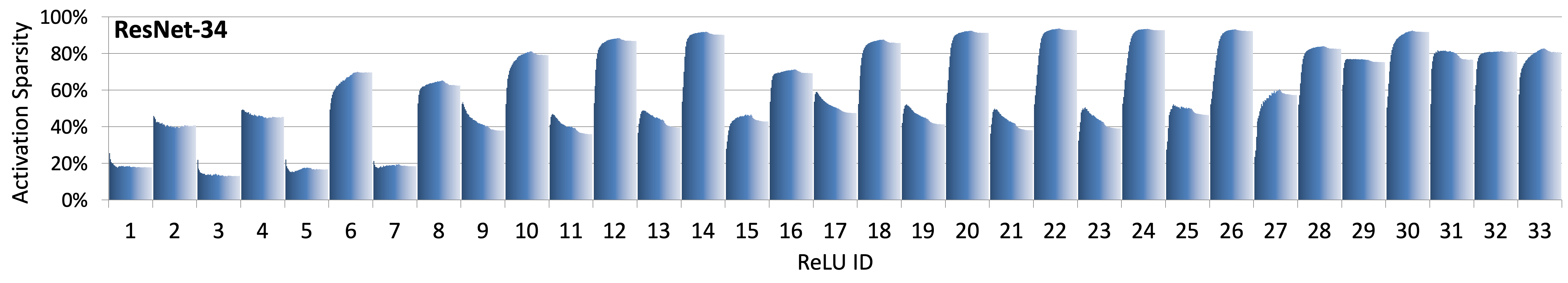}\\
    \includegraphics[width=1\linewidth]{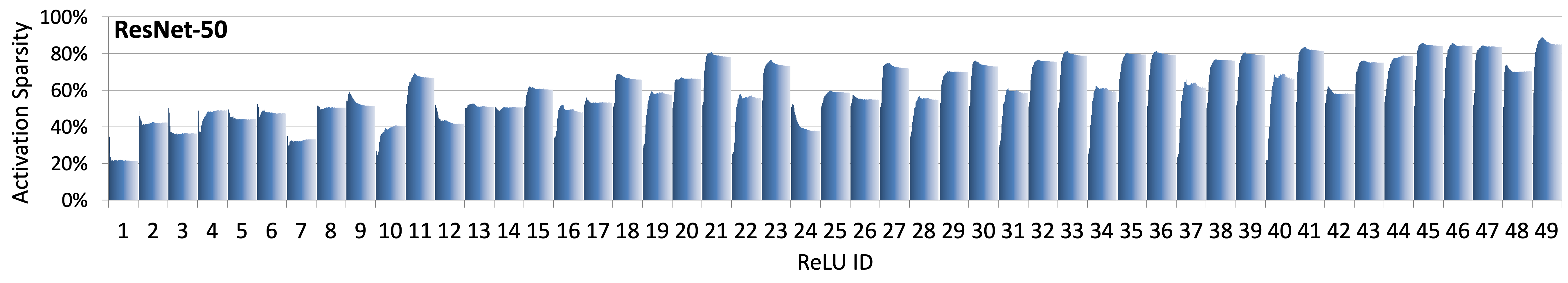}\\
    \includegraphics[width=1\linewidth]{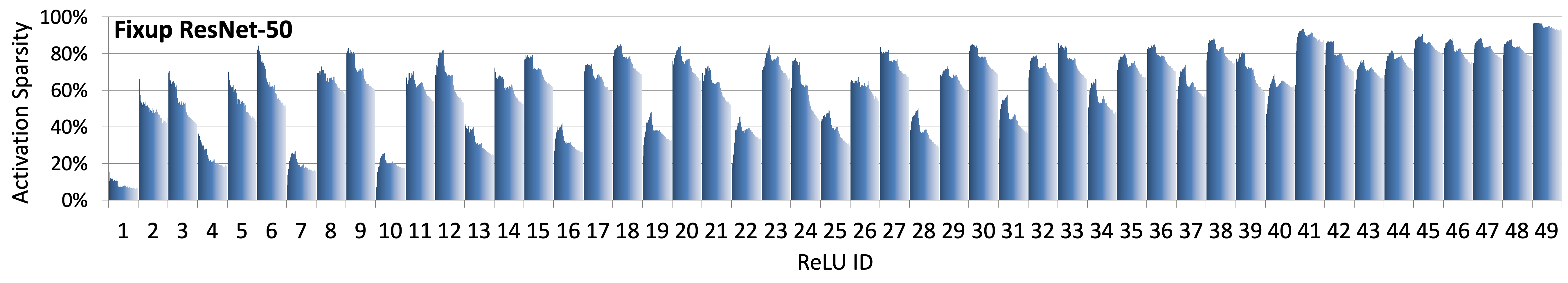}
    \caption{Measured sparsity in ReLU outputs during 100-epoch trainings on ImageNet. Each segment of the x-axis corresponds to a single layer.  Within a segment, from left to right are the sparsity from the first epoch to the last.}
    \label{fig:sparsity}
\end{figure}

Figure~\ref{fig:sparsity} presents the sparsity of each ReLU's output during our training of the three ResNet Variants. The average sparsity of each layer typically ranges from 20\% to 90\%, and the observations from Rhu et al. generally hold. One exception is that the degree of sparsity between adjacent layers fluctuates periodically; this is caused by the shortcut in each residual block, which adds positive bias to the outputs of a block and lowers the sparsity from the subsequent ReLU. The fluctuation is more pronounced in ResNet-34 and Fixup ResNet-50 than in ResNet-50.


The inclusion of BatchNorm affects \emph{\name}'s execution time of both BWI and BWW. Because ResNet-34/50 has BatchNorm, $\partial L/\partial Y$ has no sparsity, so we replace \emph{\name} with the baseline for BWI, and use the sparsity pattern in $D$ to measure the execution time of BWW. On the other hand, VGG-16 and Fixup ResNet-50 do not have BatchNorm, so we use the sparsity pattern in $\partial L/\partial Y$ to measure the execution time of BWI, and choose the higher average sparsity from $D$ or $\partial L/\partial Y$ to measure the execution time of BWW.

\begin{figure}[htbp]
    \centering
    \includegraphics[width=\linewidth]{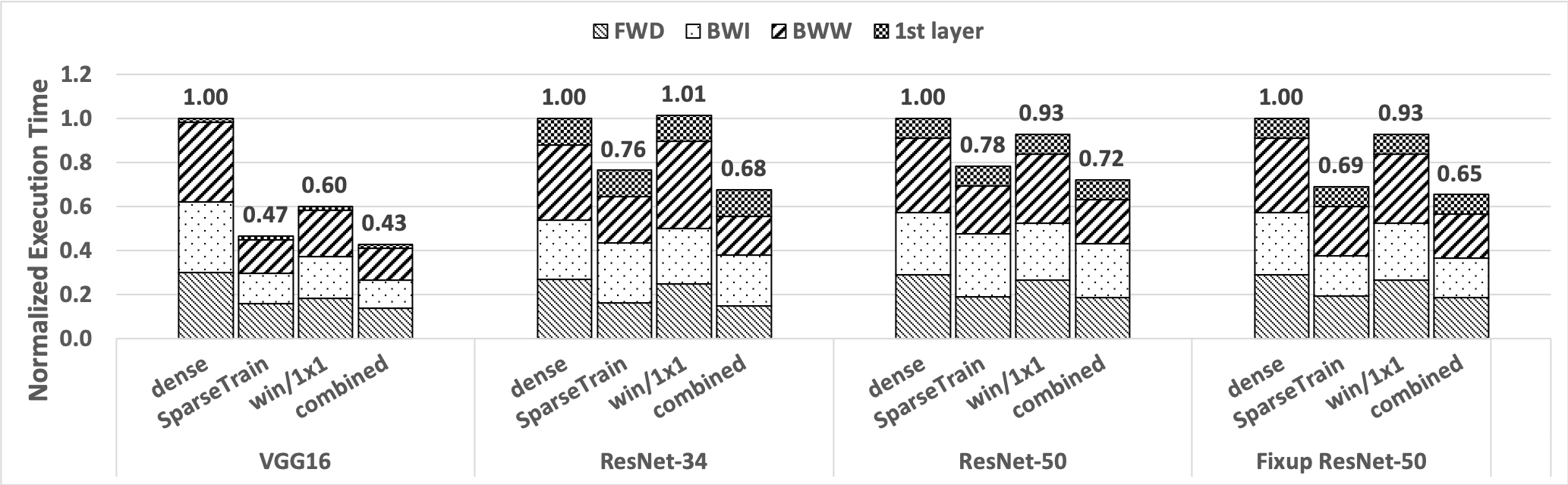}
    \caption{Breakdown of the estimated execution time of all conv layers from networks during end-to-end training, normalized to \emph{direct}}
    \label{fig:exetime}
\end{figure}

Figure~\ref{fig:exetime} illustrates the estimated total execution time of the conv layers with different algorithms during end-to-end training, normalized to the execution time of \emph{direct}. The plot stacks the execution time of each component. Because \emph{\name} is not applicable to the first layers in the network due to the input images often being zero-free, we show the execution time of the first layer as a constant overhead. 

In the plot, the \emph{\name} bars are the execution times of using purely \emph{\name}, or in the case of the ResNet-34/50, \emph{\name} for FWD and BWW plus the baseline for BWI. The \emph{win/1x1} bars are the execution times of using the \emph{Winograd} convolution or the \emph{1x1} kernel whenever possible. Because we found that \emph{\name} and \emph{Winograd} may complement each other, we also include the \emph{combined} bars that contain the execution times with the preferred convolution implementation of each layer being employed. Because the \emph{im2col} implementation is much slower than dense direct, we omit it in the plot.

Table~\ref{tab:finalspeedup} lists the speedup on the conv layers both including and excluding the first layer. The results suggest that when including the first layer, \emph{\name} speeds up the training of the conv layers in the studied networks by 1.28x-2.15x. By choosing the best algorithm for each layer, we can speed up training by 1.39x-2.35x. Note that \emph{combined} chooses the algorithm for each layer statically according to the average execution time. If we profile the sparsity of each layer at intervals during training and then dynamically select the best implementation to use based on the current sparsity level, the potential speedup may be higher.

\begin{table}[htbp]
  \caption{Projected speedup on all conv layers from networks}
  \label{tab:finalspeedup}
  \footnotesize
  \begin{tabular}{l|p{0.15\linewidth}p{0.1\linewidth}p{0.08\linewidth}|p{0.15\linewidth}p{0.1\linewidth}p{0.08\linewidth}}
    \toprule
     & \multicolumn{3}{c|}{Incl. 1st layer} & \multicolumn{3}{c}{excl. 1st layer}\\
     & \name & win/1x1 & comb. & \name & win/1x1 & comb.\\
    \midrule
    VGG16 & 2.15 & 1.66 & 2.35 & 2.19 & 1.68 & 2.40\\
    ResNet-34 & 1.31 & 0.99 & 1.48 & 1.37 & 0.98 & 1.58\\
    ResNet-50 & 1.28 & 1.08 & 1.39 & 1.31 & 1.09 & 1.44\\
    Fixup ResNet-50 & 1.45 & 1.08 & 1.53 & 1.51 & 1.09 & 1.62\\

  \bottomrule
\end{tabular}
\end{table}

\emph{\name} can speedup Fixup ResNet-50 by 1.45x instead of 1.28x on the original ResNet-50 thanks to the absence of BatchNorm. We also experimented with minibatch $N=\{32,64\}$, and confirmed that \emph{\name}'s execution time scales linearly with $N$.


\subsection{Limitations}\label{sec:limit}

Apart from the complications caused by BatchNorm, several other factors may limit the application and/or performance of \emph{\name}. First, \emph{\name} is inapplicable to networks that use activation functions other than ReLU. Nonetheless, ReLU is by far the most popular activation function for CNN.


Second, although we applied Algorithm~\ref{al:branch} to combat branch misprediction, the misprediction rate is still noticeable due to the low trip count of the transformed loop ($\leq{}V$). Further reducing mispredictions in software may be hard; however, because the trip count is generated outside of the loop body, previous hardware proposals~\cite{cfd2} can remove the branch misprediction entirely by decoupling trip count generation and loop execution.

Third, the sparsity in FWD and BWI is fully exploited because only one of the two source operands in their FMAs contains sparsity. However, both FMA operands in BWW may be sparse, so the sparsity in BWW is not fully leveraged. Also, \emph{\name} does not take advantage of the sparsity in the weights if they are iteratively pruned during training. 

Finally, due to the vectorization of the zero-checking in BWW being along the minibatch dimension, we require the batch size to be a multiple of $V$ for maximum performance.


%% file: relatedworks.tex
\section{Related Works}\label{sec:relatedworks}

Various works compress DNN models by eliminating redundant weights. Network pruning ~\cite{DeepCompression} \cite{park2016faster} removes redundant network
connections under reasonable criteria. Weight quantization~\cite{INQ}\cite{TTQ} sacrifices numerical precision to reduce model size. Wen et al.~\cite{structured-sparsity-lasso} studies structured sparsity. Their compressed models are more hardware-friendly. However, they are not applicable during training and do not exploit dynamic sparsity in the activation.

meProp~\cite{sun17meprop} sparsifies the back propagation of LSTMs and MLPs by only propagating a small number of gradients in each pass. This reduces back propagation time for the studied networks and lowers overfitting as a byproduct. Yet, it does not affect the forward propagation, nor has it been tested on CNNs. Our work is orthogonal to it and can potentially be applied in conjunction with it.

Several hardware proposals targeting DNN accelerators exploit the sparsity in weights, activations, or both. Cnvlutin~\cite{Cnvlutin} leverages sparsity in activations to skip ineffectual computations. Eyeriss~\cite{Eyeriss} clock-gates computation path and local buffer when zero is detected in the activation, and it performs convolution in a synchronous manner, collecting partial results from neighbor processing elements, therefore saving energy. However, cycles are not saved. Cambricon-X~\cite{Cambriconx} focuses on weights sparsity and skips multiplications associated with zero weights obtained by pruning. EIE~\cite{EIE} exploits the sparsity in both weights and activations using a compressed representation, but it is limited to matrix-vector multiplication (e.g. fully connected layers) and cannot accelerate the most time-consuming convolutional layers. SCNN~\cite{SCNN} utilizes the sparsity in both weights and activations, and accelerates the convolution layer. These accelerators modifies the hardware structure while our work is software only.


Normally, the Winograd algorithm~\cite{Winograd} erase the dynamic sparsity in the activation. Liu et al.~\cite{Liu2017EfficientSC} restore the activation sparsity by applying ReLU to the activation after transforming to the Winograd space. However, their approach changes the network structure. In addition, their focus is to reduce the operation count for running DNN inference on mobile devices, and they do not target training nor efficient vectorized implementation.

%% file: conclusion.tex
\section{Conclusion}\label{sec:conclusion}

The widespread usage of the ReLU non-linear activation function in DNNs means that DNN training includes a significant fraction of computations on zero values.  Traditional sparse methods, however, are not effective since the fraction of zeros is modest, and the locations of zeros are dynamic.

We observe that each output value from a ReLU function sees significant reuse in all three phases of training.  Therefore, if we order the main compute loops appropriately, we can check for zero on the fly, and potentially jump over chunks of work. We further vectorize the sparsity-checking, maximize efficiency when sparsity levels are low, and minimize branch mispredictions.  When applied to direct convolutions, at $0\%$ sparsity, our approach generally performs within $10\%$ of a highly optimized dense code. For training of real DNNs, our approach is projected to outperform the dense convolutions by 1.31x-2.19x.


This paper is the first work to exploit dynamic sparsity with only software techniques and opens up new research direction in speeding up computation with modest sparsity.